\documentclass[11pt]{article}
\usepackage[preprint]{acl}
\usepackage{times}
\usepackage{latexsym}
\usepackage[T1]{fontenc}
\usepackage[utf8]{inputenc}
\usepackage{microtype}
\usepackage{inconsolata}
\usepackage{booktabs}
\usepackage{float}
\usepackage{amsmath}
\usepackage{graphicx}
\graphicspath{{figs/}}

\title{Inherited Heads\\
\large Audio language models track speakers with their text backbone's attention,
and an attention-mass ranking retrieves a different set}
\author{Bojro Das \\
  Cornell University \\
  \texttt{bd463@cornell.edu}}
\date{}

\begin{document}
\maketitle

\begin{abstract}
Asked to describe what one of six speakers in a recording talks about, audio
language models describe the right one on 6 to 16\% of trials, below the 16.7\% a guess would give.
Adding a fixed bias to the attention logits of a hundred heads, under a tenth of
the model's and with no training, redirects the description to whichever speaker
we choose, on 90.7\% to 99.0\% of trials.

Those heads are largely not specific to audio. Rank the text-only language
model an audio model was built from, or a released model of the same family, on a
written version of the task, take its top hundred heads,
and carry them over unchanged: they redirect the audio model on 80.8\% to 95.0\%
of trials, with nothing about audio entering the selection. The audio and text
head sets share 66 to 74 of 100 where chance would give about 20, and the shared
part alone reproduces almost all of the steering. What that does not show is that
sharing is what makes the heads work: an equal-sized draw from the same
discovered hundred does nearly as well, and none of our three models separates
the two explanations.

A second finding concerns how such heads are found. Ranking heads by how much
attention they place on the segment asked about, as an established score does, or
by how much of their attention moves with the question, as a per-head normalised
variant does, gives top hundreds that share 69, 37 and 4 heads across our three
models. In Ultravox, where they share 4, the established score's heads leave
output the judge cannot place on any segment on 69.7\% of trials, against 40.0\%
with no intervention and 1.0\% for the normalised variant. That is one arm of
six; on the other five the established score steers above a random draw.
\end{abstract}

\begin{figure*}[t]\centering
\includegraphics[width=\textwidth]{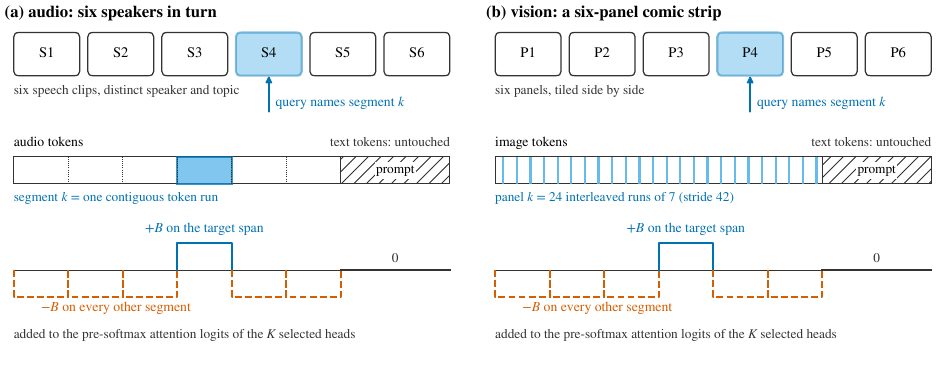}
\caption{The task and the edit. Six speakers read in turn in one 33.5-second
strip; the model is asked to describe what one of them talks about. The
intervention adds a constant $+B$ to the pre-softmax attention logits of the
selected heads at every key inside the target segment's token span and $-B$
inside the other five, in prefill and during decoding. It is positional and
content-independent: no learned direction, no second prompt.}
\label{fig:task}
\end{figure*}

\section{Introduction}
\label{sec:intro}

Six people speak in turn in a single recording. Ask an audio language model to
describe what one of them talks about and it names the speaker we asked for on 6
to 16\% of trials, below the 16.7\% a guess would give, because it often
describes the wrong speaker or no identifiable one. The failure is not particular to this
testbed: a benchmark of twenty speech language models finds they attribute speech
to the wrong speaker and fall back on semantic priors rather than vocal cues
\citep{hear2026}. Add a constant to the attention logits
of a hundred heads, under a tenth of the model's, with no training and no labels,
and it describes whichever speaker we choose on 90.7\% to 99.0\% of trials.
Figure~\ref{fig:task} shows the task and the edit.

\textbf{The machinery an audio model uses to track who is speaking is machinery
its language backbone already had.} The sharpest evidence needs no audio at all.
Rank each model's text-only backbone, or for Qwen2-Audio a released model of the
same family, on a written version of the task, take
its top hundred heads, and carry those indices over unchanged: they steer the
audio model to a chosen speaker on 95.0\%, 85.0\% and 80.8\% of trials, against
12.5\%, 45.0\% and 15.0\% for a single band-matched random hundred. Nothing about audio entered that selection. What transfers is a set of
head indices rather than a direction in activation space, and the model they were
ranked on has no audio pathway at all.

The head sets agree. Ranked on audio and on text, the two share 66, 71 and 74 of
100 indices against layer-matched chance overlaps near 20, and the shared part is
not merely present but sufficient: used alone it recovers almost all of the
steering, and on one model it is statistically indistinguishable from the full
set.

Sufficiency is a claim about that subset and we are careful not to stretch it. A
random draw of the same size from the same discovered hundred also recovers most
of the behaviour, so what the intervention establishes is that two thirds of this
head set carries the effect, not that being shared with the backbone is what
makes those heads work. None of the three models separates those two
explanations; the strongest case, on Qwen2-Audio, is an ordering three random
draws would produce by chance a quarter of the time.

A second finding concerns how such heads are found. The established way ranks
heads by how much of their attention lands on the region asked about. How much a
head attends there and how much of its attention moves with the question are only
weakly related in every audio model we test, and the two rankings share 69 of their top
100 heads in Qwen2-Audio, 37 in SALMONN and 4 in Ultravox. On Ultravox the
mass-ranked heads steer at 1.7\% against the selectivity ranking's 99.0\%, and the
two sets fail in different ways: under the mass-ranked heads the judge cannot
place the description on any segment on 69.7\% of trials, against 40.0\% with no
intervention and 1.0\% under ours. This is one arm of six, and on the other five
the mass ranking steers above a band-matched random draw.

\paragraph{Contributions.}
\begin{itemize}\itemsep2pt \parskip0pt
\item \textbf{Sufficiency, not co-location.} Three audio language models analysed
at head level against their text backbones, with the shared subset shown
causally sufficient for segment tracking under intervention, and with heads
chosen on the text model alone shown to steer the audio one. Whether
\emph{sharedness} is what matters, rather than membership of the discovered set,
is not separated on any of the three, and reported as such.
\item \textbf{Two rankings of the same attention retrieve partly different head
sets} in every audio model tested, sharing 69, 37 and 4 of their top 100 heads.
The tracking heads' median sits at or above the typical head in attention to the
audio, and most of them sit below a top-100 cut on it.
\item \textbf{A published head-selection score whose heads, under intervention,
mostly leave the output unattributable rather than misdirecting it}: 69.7\% of
trials, against 40.0\% with no intervention and 1.0\% for a per-head normalised
variant of the same score, which reaches near ceiling. On one arm of six, with
the two head sets about seventeen layers apart, so without a control that separates
the score from depth.
\item \textbf{Two measurements of our own instruments}, both of which we would
have preferred not to find. At the strength where our intervention saturates,
four fifths of the attention the target gains comes from prompt text and the
attention sink rather than from the other segments. And layer-matched random head
controls at these set sizes are not point estimates: five draws span 0.642 of
accuracy on one model against 0.025 on another, enough that a difference quoted
against a single draw can carry the wrong sign.
\end{itemize}

\section{A family of scores, at full strength}
\label{sec:family}

The scores this paper examines share a shape: run the model once per candidate
region, record how much attention each head places on each region, and rank heads
by the mass on the region that was asked about. The family is attractive for good
reasons. It needs no labels, no training, and no gradients, only a handful of
forward passes, and it returns a small head set that can be acted on directly.
The instance we test against is the gaze score of \citet{gandikota2026gaze},
which ranks heads by the mass they place on the queried region; we call the
resulting order \emph{the mass ranking} throughout.

The mass ranking mostly works. In the setting it was introduced for it recovers
head sets that steer a vision-language model's answers to a chosen panel at
83.1\% accuracy. In our own runs the mass ranking steers above a band-matched
random draw on five of the six arms we measure: 0.647 against 0.060 on
Qwen2-Audio, 0.807 against 0.170 on SALMONN, 0.175 against 0.067 on Qwen2-VL,
where it is itself close to chance, and 0.408 against 0.225 and 0.383 against
0.042 on the two Bunny layouts. On the sixth arm it does not, and
Section~\ref{sec:critique} takes that up. Each of those controls is a single draw
over the layers the selectivity-ranked hundred spans, which is weaker evidence
than Section~\ref{sec:cost} would ask for.

Nor is it alone. A related raw attention-mass criterion in multimodal models
finds heads whose masking degrades long-context performance more than masking a
random set \citep{core2026heads}, though their control is matched in number
rather than in layer and their test is ablation rather than steering. Path
patching across five multimodal models likewise isolates head groups that beat a
size-matched random control under ablation, averaged over five draws
\citep{modalityconflict2026}. The
family already has audio members, and they go further than a scalar. Heads in an
audio language model can be scored by how much of the final prompt token's
attention lands on the audio at all, kept where that mass correlates with
answering correctly, and then steered through their outputs to raise
multiple-choice accuracy on Qwen2-Audio, against a same-size random-head control
\citep{audiospecialistheads}. Two more recent works give that quantity a
coordinate axis, scoring each head by whether the audio it attends to most lines
up in time with the word being generated, and using the result to decide whose
audio cache to keep \citep{audiokv2026, wnw2026}. Head-level span tracking in
audio language models is therefore not something we introduce, and neither is a
random-head control for steering through selected heads. What none of the three
does is edit the heads' attention itself, and none compares a head set against
the model's text-only backbone. In audio
specifically, activation steering has been used to redistribute a model's
attention over time and improve sound-event localisation against a random
baseline \citep{steeringwheretolisten}, on Qwen2-Audio and Audio Flamingo 3. That
intervention adds a vector to the residual stream rather than acting on selected
heads, where ours biases attention logits on a selected subset of them. Two audio papers come
closer to our recipe than to our task. One scores the heads of a decoder-only
text-to-speech model and then constrains attention on the selected ones at
inference \citep{alignmentmaps2024}, which is our two-step shape exactly, aimed
at alignment robustness rather than at choosing a speaker. The other masks heads
in an audio model's own text backbone \citep{ahamask2026}, with masks trained
per task rather than applied positionally, and with no comparison against that
backbone's own text behaviour.

The method's authors are also careful in ways a critique should not obscure. They
motivate ranking by mass rather than by a contrast explicitly, as a guard against
heads that appear well aimed by accident; their judge abstains rather than
guessing and counts abstentions as misses against a fixed denominator; and they
hedge their causal reading in the text rather than in a footnote.

What the family does not do is separate two things its central quantity
conflates. A head can place a large share of its attention on the modality
because it is tracking what was asked, or because it places a large share there
regardless of what was asked. Ranking by the raw share scores both alike. The
remedy, normalising per head and asking what fraction of a head's modality
attention moves with the question, is not ours. Relative and contrastive head
criteria are established in vision, both as a small set of grounding heads
selected by contrast \citep{contrastiveheads2026} and as a
visual non-sink ratio \citep{kang2025sink}, and in text as a logit-contribution
score \citep{locos2026}. Nor is the observation that raw attention is a
poor proxy for causal importance, which goes back at least to norm-weighted
attention \citep{kobayashi2020attention} and has been argued at head level across
twelve scoring functions \citep{inactiveheads2026}. The random-baseline comparison
we put both variants through is likewise one test in a published protocol
\citep{unittests2026}.

We contribute no score and no control. What we contribute is the result of
putting the two variants of one quantity through the same causal test, at head
level, in a modality where that has not been done. They retrieve partly different
head sets in all three audio models, sharing between 69 and 4 of their top 100
heads, and in the model where they diverge most the two sets fail in different
ways under the same intervention, one leaving output the judge cannot place on
69.7\% of trials while the other reaches ceiling.

\section{Method}
\label{sec:method}

\paragraph{Task and data.} Each item is a 33.5-second strip of six five-second
LibriSpeech segments read by six different speakers, separated by silence. The
model is asked, in one neutral prompt, to describe what one speaker talks about.
The corpus is 500 such strips. Discovery uses items 0--49 and the confirmatory
steering results are measured on held-out items 100 onwards. A confirmatory cell
is 50 strips by 6 target segments, or 300 trials; the subset and control
comparisons in Sections~\ref{sec:borrowed} and~\ref{sec:cost} use 20 strips, or
120 trials, and are compared only against each other. Two sets of numbers are not on the
held-out split and say so where they appear: the size sweep in
Section~\ref{sec:borrowed} and the text-derived transfer in the same section are
both on items 0--19, which is inside the discovery split.

\paragraph{Judge.} A lexical judge scores which segment a description matches by
transcript overlap, and abstains when no segment matches. Abstentions count as
misses against a fixed denominator, so chance remains $1/6$. We call the share of
trials on which it abstains the \emph{unattributable rate}, and report it beside
accuracy wherever the two come apart: a head set that makes the model describe the
wrong speaker and one that makes it describe nobody score the same on accuracy
alone. Lexical scoring is
primary by preregistration. We also re-score the stored per-segment outputs under
a forced choice, which raises accuracy by removing abstentions, and report it as
a secondary analysis; the abstention rate is material on the vision arms (27.5\%
on Qwen2-VL) and small on audio.

\paragraph{Models.} Qwen2-Audio-7B-Instruct \citep{qwen2audio2024}, whose
language model starts from Qwen-7B and is trained jointly with the audio encoder;
Ultravox-v0.6-8B,\footnote{\texttt{fixie-ai/ultravox-v0\_6-llama-3\_1-8b}} built on
a frozen Llama-3.1-8B-Instruct; and SALMONN-13B \citep{salmonn2024}, built on a
frozen Vicuna-13B adapted with LoRA, whose LoRA and Q-Former together train about
0.24\% of parameters. Together they span the range
of how much an audio model's text backbone is altered. For Ultravox and SALMONN the
text model we rank is the backbone itself. For Qwen2-Audio no text checkpoint of
the jointly trained language model exists, so we rank Qwen1.5-7B-Chat, a released
model of the same family with an identical attention layout; that comparison is a
match of lineage and architecture rather than of weights. The first two have 32 layers of
32 attention heads, or 1024 in total; SALMONN has 40 of 40, or 1600. We write
$K$ for the number of heads a set contains, so the $K{=}100$ used throughout is
under a tenth of the model in every case. All three run in 4-bit NF4 with
bf16 compute, which is a real threat to head-ranking fidelity and is addressed in
Limitations.

\paragraph{Scores.} Ask the model about each segment in turn and record, per head,
the share of the final prompt token's attention landing on each segment's token
span, giving a $6\times6$ matrix. The published score is the mean of its diagonal, which
is the share of the head's attention that lands on the queried segment. We call
that quantity the head's \emph{mass}. Our variant divides each row by its total
before averaging, so the score is the fraction of a head's modality attention
that moves with the question rather than the absolute amount; we call that its
\emph{selectivity}. Throughout, \emph{the mass ranking} and \emph{the
selectivity ranking} mean heads ordered by one quantity or the other. A head with
high mass puts a lot of attention on the segment asked about, however much it puts
on the rest of the audio. A head with high selectivity puts most of its audio
attention on whichever segment was asked about. The two are
different questions about the same attention, and the rest of the paper uses both
names. Both are computed from the same forward passes. A
\emph{real-versus-shuffled} statistic of 1.0 means a ranking is indistinguishable
from one computed with the segment labels shuffled. One further term recurs: the
\emph{attention sink} is the early position, here the beginning-of-sequence
token, on which these models park large amounts of attention that carries no
information about the input. Scores are read at the final prompt token, which is
the point the published method uses; we show in Section~\ref{sec:null} that this
choice understates head structure.

\paragraph{Intervention.} To steer towards segment $t$ we add $+B$ to the
pre-softmax attention logits of the selected heads at every key inside segment
$t$'s token span, and $-B$ at every key inside the other five segments' spans;
prompt tokens, the BOS position and all other keys are left untouched, in prefill
and during decoding. The edit is positional and content-independent, defined
by token spans, with no learned direction and no second prompt. Because the bias is added before the softmax, its size decides how sharp the
resulting distribution is: at $B{=}10^4$ a selected head attends to almost
nothing but the target span, while at $B{=}5$ the shift is mild and the head's
other attention survives. We report $B{=}10^4$ for comparability with published
work and $B{=}5$, where the effect already saturates, for everything that depends
on operating at a strength the model might plausibly reach on its own. The same
kind of additive bias, applied at every head of the hooked layers rather than to
a selected head set, has been used to redirect a driving model's attention toward
safety-critical actors \citep{prasad2026steering}; their per-call audit found the
bias silently never reaching one pathway, which is why every run here asserts that the hooked and
unhooked attention differ before its numbers are used.

\paragraph{Controls.} Head sets are read against random head sets, and two
choices define one: how the draw is matched to the set it controls, and which
pool it is drawn from.

Matching comes in two strengths. A \emph{band-matched} draw is taken uniformly
from the range of layers the selectivity-ranked hundred spans, excluding that
hundred, and the same draw is read beside both rankings on a model even though it
spans the selectivity ranking's layers rather than the mass ranking's
(Section~\ref{sec:critique}). A \emph{layer-matched} draw is the stronger one,
reproducing a given set's per-layer counts exactly, so it has the same number of
heads in each individual layer rather than merely the same span; we name the set
it copies wherever one appears. The two do not agree. On Ultravox, the one model
with both at $K{=}100$, both band-matched draws sit above eight of the ten
layer-matched ones, so band-matched figures are marked as such.

The pool is usually everything outside the discovered top-100, so that a
difference cannot be a redraw of the set being tested. Section~\ref{sec:borrowed}
needs one exception and states it there: to ask whether it matters that a subset
is shared with the text backbone, rather than that it is in the discovered set at
all, the draw has to come from \emph{inside} that set. Those draws are matched on
size only, taken uniformly from the discovered hundred. The controls behind our
subset comparisons, and the layer-matched controls at $K{=}100$ on Ultravox, are
reported over three to ten draws rather than as single numbers, with every draw
in Appendix~\ref{app:sets}; other random controls in the text are single draws
and are marked as band-matched where they are. We quote no bootstrap interval on
a difference
between a fixed set and a random control; Section~\ref{sec:cost} gives the
measurement behind both practices. Intervals appear only between two fixed
sets on the same items, resampling strips rather than trials, 10{,}000 resamples.
Neither the random-baseline test nor the matching is ours: a same-budget random
baseline is one of a published set of unit tests for interpretability claims
\citep{unittests2026}, and size- and layer-matched random draws have been used as
controls for circuit claims before \citep{shengfu2026matched}.

\paragraph{What reliability does not buy.} On Ultravox, the one model where we
measured it, the ranking recovers 94 of its top 100 heads on disjoint held-out
items. That is measurement precision only. Stability
without causal specificity is not sufficient evidence of a task-causal
mechanism, as has been shown for crosscoder features \citep{seedstability2026}, and no claim here rests on it.

\begin{figure*}[t]\centering
\includegraphics[width=\textwidth]{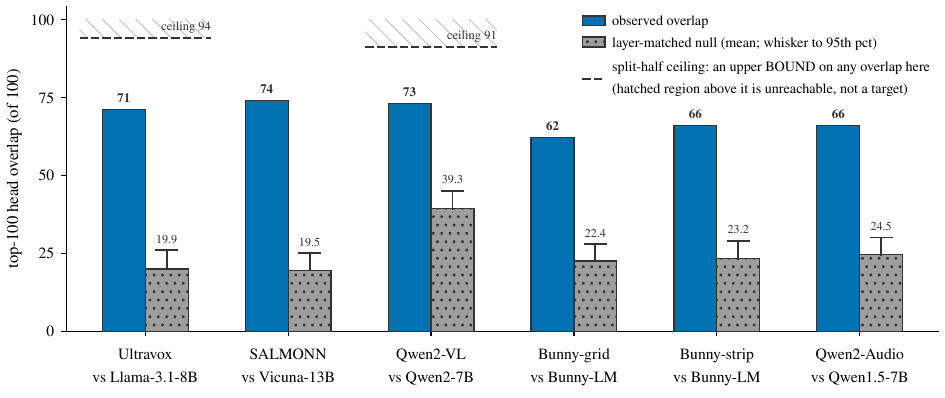}
\caption{Head-set overlap between each multimodal model and its text-only
backbone (for Qwen2-Audio, a released relative with the same layout), against a null that resamples 100 heads with the same layer histogram.
The null, not the naive $k^2/N$ baseline, is the right comparison because these
heads concentrate in the middle layers. Split-half ceilings are drawn where a
reliability run exists; where one does not, no bound is drawn rather than a
stand-in.}
\label{fig:inheritance}
\end{figure*}

\begin{figure*}[t]\centering
\includegraphics[width=\textwidth]{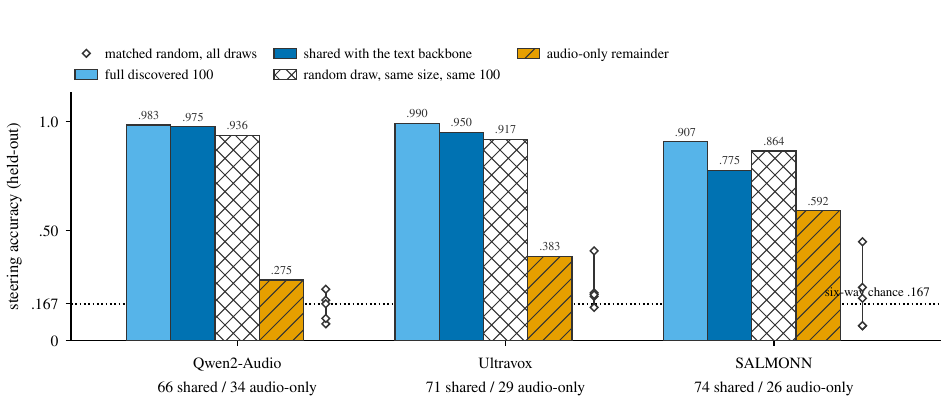}
\caption{Each subset steering alone. The third bar is the one that constrains
the reading: a random draw of the shared subset's size, taken from the same
discovered hundred, reaches within 0.039 of the shared subset on two models and
exceeds it on SALMONN. Any two-thirds of this head set carries most of the
behaviour, so sufficiency is a claim about the subset rather than about
sharedness. Diamonds are the five draws of a layer-matched random control for
the shared subset, taken from outside the discovered hundred; the shared subset
exceeds every one of them on every arm.}
\label{fig:sufficiency}
\end{figure*}

\section{Tracking runs on inherited heads}
\label{sec:borrowed}

Choose a hundred attention heads without ever looking at audio. Give the
text-only model an audio model was built on (for Qwen2-Audio, a released model of
the same family; Section~\ref{sec:method}) the written version of the same task, six
transcribed passages labelled \emph{Segment 1} to \emph{Segment 6} with a
question about one of them, rank its heads, and carry the top hundred indices
over to the audio model unchanged. That head set then steers the audio model to a
chosen speaker on 95.0\% of trials in Qwen2-Audio, 85.0\% in Ultravox, and
80.8\% in SALMONN, against 12.5\%, 45.0\%, and 15.0\% respectively for a single
band-matched random hundred on each. Nothing about audio entered the
selection.

That is the cleanest evidence in this paper, because it is the only test here
whose head set was chosen without touching the modality it is used on. It also
separates two explanations the rest of the section cannot. On Ultravox the
backbone is frozen, so shared heads are also shared weights and the transfer
could be nothing more than that. On Qwen2-Audio the language model was trained
jointly with the audio model, and the text model we rank is a same-family model
rather than its starting point, so the weights differ twice over, and the
text-derived heads
still come within 0.008 of the heads chosen on audio directly. That is consistent with the position mattering more than the weights, though a
shared layout alone does not guarantee that head indices correspond across
checkpoints. The machinery an audio model uses
to track who is speaking is machinery its language backbone already had.

One caveat attaches to the numbers in the last two paragraphs. They are measured on the first twenty
strips, which sit inside the split the audio head sets were discovered on rather
than the held-out split used everywhere else here. That cuts both ways, and the
net is conservative: the text-derived rankings never saw an audio strip, so for
them these items are effectively held out, while the audio-derived comparison is
in-sample and flattered.

The head sets bear that out. When we rank each model's text model on
the written task, its top hundred and the audio top hundred share 66 head indices
in Qwen2-Audio, 71 in Ultravox, and 74 in SALMONN. To know whether that
is a lot, we ask how many heads two such sets would share by chance if one were
drawn at random from the same layers the real set occupies: about 20 in every
case (24.5, 19.9, and 19.5, with 95th percentiles of 30, 26, and 25).
Figure~\ref{fig:inheritance} plots both against that null. Matching
the layers matters, because these heads cluster in the middle of the network and
a baseline that ignored depth would be far too easy to beat.

Two things qualify that count. Each comparison is against a text model with the
same attention layout, so a shared index is the same position in both; whether it is the same
\emph{parameter} depends on the model. Ultravox's backbone is frozen, so there it
is literally the same tensor. SALMONN's is LoRA-adapted, and Qwen2-Audio's is
trained jointly with the audio model and compared against a same-family model
rather than its starting checkpoint, so on those two the weights differ and the claim is
positional. And the ranking is not perfectly repeatable: rediscovered on a disjoint set of
audio clips it returns 94 of the same 100 heads on Ultravox, the one arm where we
measured this, so of that model's 71 shared heads about four are expected to be
ranking noise rather than shared structure.

Overlap of this kind is not new and we do not claim it
\citep{li2026intrinsic,choi2026truth,park2026vrh,bektursun2026borrowed}, and
\citet{cascade2026} makes matched-backbone testing its method. What an overlap
count cannot establish is what the shared heads \emph{do}. In a controlled setting, training pins down only what a group of heads
contributes in total, not how that total is divided among them
\citep{computationalunit2026}, so two models can use the same heads and still
split the work between them differently. The way past that is to intervene on the parts separately.

We split each discovered top-100 into the heads that are also in its backbone's
text top-100 and the heads that are not, then steer with each part alone on the
same held-out strips.

\paragraph{Read each part against a draw of its own size.}
Set size is not neutral here. Our own ranking steers at 0.283 with its top 25
heads, 0.692 with 50, and 0.983 with 100; that sweep is on the discovery clips,
so its values are not comparable with the table below, but its shape is what
matters. A subset of 29 heads is expected to score below one of 71 whatever the
two contain, so comparing the shared part against the unshared part directly
would measure size as much as sharedness, and we do not do it. Instead each part is read against a control built like this:
take the same discovered hundred, and sample from it at random as many heads as
the part being tested has. So the shared part of Ultravox, 71 heads, is read
against 71 heads drawn at random from the same 100, and its unshared part, 29
heads, against 29 drawn the same way. Table~\ref{tab:sufficiency} reports every
cell that way, and Appendix~\ref{app:sets} gives the individual draws.

\begin{table*}[t]\centering\small\setlength{\tabcolsep}{3.5pt}
\begin{tabular}{lccccc}
\toprule
 & & \multicolumn{2}{c}{shared with the backbone} & \multicolumn{2}{c}{not shared} \\
\cmidrule(lr){3-4}\cmidrule(lr){5-6}
model & all 100 & these heads & random, same count, from the 100 & these heads & random, same count, from the 100 \\
\midrule
Qwen2-Audio & 0.983 & 0.975 \,(66) & 0.936 & 0.275 \,(34) & \textbf{0.644} \\
Ultravox    & 0.990 & 0.950 \,(71) & 0.917 & 0.383 \,(29) & 0.444 \\
SALMONN     & 0.907 & 0.775 \,(74) & 0.864 & 0.592 \,(26) & 0.511 \\
\bottomrule
\end{tabular}
\caption{Each part of the head set steering on its own, against a random draw of
the same size from the same discovered hundred. Head counts in parentheses; same
held-out strips throughout; chance is $1/6$; 20 strips $\times$ 6 targets $=$ 120
trials per cell. Each random figure is the mean of three random draws of the
same size taken from inside the discovered hundred, the exception to the usual pool noted
in Section~\ref{sec:method}. Per-draw values are in
Appendix~\ref{app:sets}.}
\label{tab:sufficiency}
\end{table*}

\paragraph{Two thirds of the head set is enough, whichever two thirds.}
On Qwen2-Audio 66 heads do what 100 do, with the difference from the full set at
$-0.008$ and its interval spanning zero; Ultravox falls short by 0.040 and
SALMONN by 0.132. A random draw of the same size reaches 0.936, 0.917, and 0.864,
within 0.039 of the shared part on two models and above it on SALMONN. So the
head set tolerates losing a third of its members, and is largely indifferent to
which third goes. Figure~\ref{fig:sufficiency} puts the shared part beside a
layer-matched control drawn from outside the hundred, with all five of its draws
shown.

What the intervention does to reach those numbers is a separate question, and not
a flattering one. Section~\ref{sec:cost} shows that at the strength where
steering saturates, most of the attention the target gains is taken from the
prompt and from the attention sink rather than from the other speakers. That
bears on how the mechanism should be described. It does not bear on which heads
carry it, because every head set here was chosen on the unedited model.

It does not tolerate losing half. On the size sweep, cutting Ultravox's set from
100 heads to 50 takes it from 0.983 to 0.692, on the same clips and with no other
change. There is a cliff somewhere between two thirds and one half, so this is
tolerance of a moderate loss rather than general redundancy.

\paragraph{How much that bears on sharedness depends on the contrast available.}
The two controls are not equally informative, and the arithmetic says why. If 71
of 100 heads are shared and we draw 71 at random from those 100, at least 42 of
them must be shared; across our arms the minimum is 48 to 65\% and the draws came
out at 65 to 73\% shared. At that size no sharedness-free control exists to be
built, so the left-hand comparison is 100\% shared against roughly 70\% shared, a
narrow contrast. At the unshared part's size there is no such floor: those heads
contain none of the backbone's set, and a same-size draw is again about 70\%
shared, giving a contrast more than twice as wide.

Read that way the two columns agree with each other within every model. Sharing
more helps on Qwen2-Audio at both contrasts ($+0.039$ narrow, $+0.369$ wide) and
on Ultravox at both ($+0.033$, $+0.061$), and hurts on SALMONN at both
($-0.089$, $-0.081$). Qwen2-Audio's wide contrast is the only one large enough to
read: its three draws reach 0.358, 0.683, and 0.892 against the unshared subset's
0.275, so every draw exceeds it. Three draws cannot make that decisive: a
subset no different from its draws would still sit below all three of them a
quarter of the time. We quote no interval on that difference. The
random side is an average over three head sets rather than a fixed one, and an
item-level bootstrap would still not carry the variance that matters here, which
is between draws; Section~\ref{sec:cost} is the reason we hold to that rule even
where it costs us. So the subset claim holds on all three models, and sharedness
helping is consistent on two, reversed on the third, and settled on none. With two contrasts per model we read the pattern as a direction, not a
dose-response.

Two further things the table does not settle. The shared part accounts for the
full set completely on jointly trained Qwen2-Audio, less so on frozen Ultravox
and least on LoRA SALMONN, an ordering that does not track how much each backbone
was trained and for which we offer no account. And the selectivity-ranked hundred
is not an optimum. On the
discovery clips SALMONN's top 250 heads steer at 0.942 against 0.892 for its top
hundred, and on the held-out strips a single 74-head draw from within that
hundred reached 0.958, above the full set. We report the head set the discovery
procedure returns, not a tuned one.

\paragraph{Against a published negative.}
\citet{bektursun2026borrowed} keep four named heads of the 192 in a six-layer
slice of a frozen text backbone, zero the rest, and find chance error on every task, concluding the
computation is distributed. We keep 66 to 74 of 1024 to 1600, mask nothing, use a
positional attention bias rather than ablation, and work in released audio models
on the modality they were trained for. The results are compatible; the
interventions are not comparable.

\paragraph{What follows from this.}
If segment tracking runs on heads the backbone already had, those heads were not
selected for audio, and how they attend to it is not what makes them the tracking
heads. The established way to find such heads ranks them by exactly that
quantity, and Section~\ref{sec:critique} measures what it retrieves: these heads
attend to audio more than most, and a mass ranking still retrieves a partly
different set.

\begin{figure*}[t]\centering
\includegraphics[width=\textwidth]{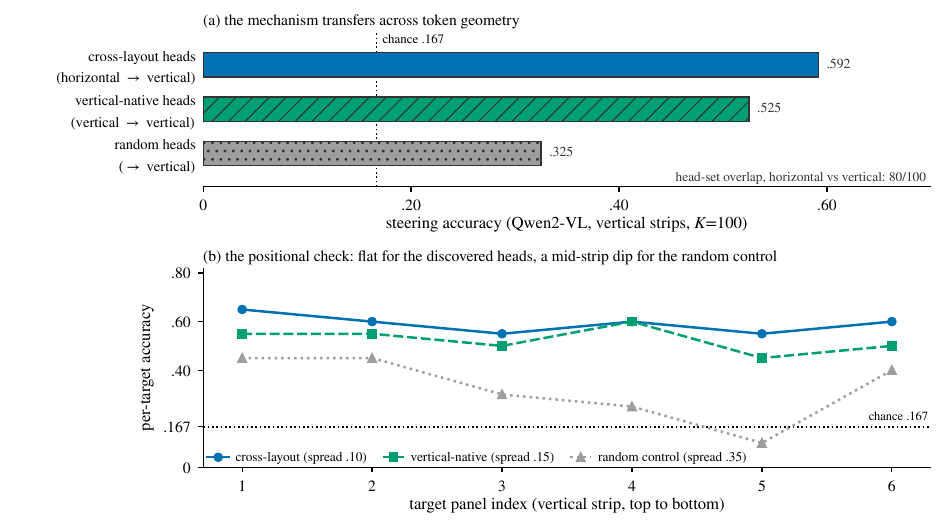}
\caption{Changing cross-modal token geometry. Patches are flattened row by row,
so tiling the panels side by side scatters each one across the sequence in 24
runs of 7 tokens, while stacking them makes each panel a single contiguous block.
Heads discovered on one layout steer the other. The two discovered sets overlap
on 80 of 100 heads, which is why we read this as suggestive rather than
decisive.}
\label{fig:layout}
\end{figure*}

\section{Does the head set survive a change of token geometry?}
\label{sec:geometry}

A head that appears to track a segment might only be tracking a position. Our
audio strips always present six segments in the same order and the same token
ranges, so a head that always attends to the fourth sixth of the input would
score well without representing anything about what was said. Separating the two requires
moving the content within the token sequence, which we did not do for audio,
where each segment's tokens are contiguous in time. We therefore run this control in the vision arm and are explicit
that it is evidence by analogy for the audio results, not evidence about them.

A vision transformer cuts an image into a grid of patches and flattens that grid
one row at a time. Tiling six panels side by side therefore interleaves them:
every row of the image crosses all six, so a single panel arrives as twenty-four
separate runs of seven tokens, each forty-two apart. Stacking the same six panels
vertically gives each panel its own rows, and it arrives as one unbroken block.
The same picture is scattered across the sequence in one layout and contiguous in
the other, which Figure~\ref{fig:layout} draws. Task, images and model are
unchanged. This is a change of cross-modal
token geometry, in which the patch grid itself differs, rather than a re-ordering
of items within one modality, and a head keyed to token positions could not find
the same panel under both.

Heads discovered on the horizontal layout steer the vertical layout at 0.592,
against 0.325 for a band-matched random control on the same layout; heads
discovered natively on the vertical layout reach 0.525, which we do not read as a
difference. The two discovered sets overlap on 80 of 100 heads, and that overlap
limits what the transfer shows: most of the transferred set is also the native
set, so this is better described as suggestive than as decisive. The experiment
that would settle it, steering the vertical layout with only the 20 heads unique
to the horizontal set, we did not run.

\paragraph{The positional confound, checked rather than assumed.}
Two mechanisms in the literature would produce apparent tracking from position
alone: over-attention to lower image regions \citep{lowerregion2025}, and
cross-modal position bias in rotary embeddings \citep{circlerope2025}. Both
predict that steering accuracy should be ordered by segment index. It is, in
every cell we measured, and what differs is by how much: per-target accuracy
spreads are 0.10 for transferred heads and 0.15 for native ones on the vertical
layout, against 0.35 for the band-matched random control on that layout and 0.35
on the horizontal one. Ultravox, the one audio model we ran this check on, is
flatter still at 0.04. Where a gradient does appear it runs the wrong way for the
flagged mechanism, which predicts more attention to later, lower regions and so
rising rather than falling accuracy with index. For audio, flatness is the whole
of the positional evidence, and it is weaker than transfer would be.

The gradient is steepest on the horizontal Qwen2-VL layout, where steering gets
worse the later the target panel sits: 0.70, 0.65, 0.65, 0.55, 0.50, and 0.35
across the six positions, a correlation between accuracy and panel index of
$-0.95$, close to a straight line down rather than noise. Steering that depends on
where the target is rather than on what is in it is the confound this section
exists to rule out, so we state plainly that on this layout we do not rule it out.

Two things bound it. It is not a property of the head set, since a band-matched
random draw declines with index as well, at $-0.53$ on the vertical layout, so
whatever produces the gradient sits in the model or the task rather than in what
the score selects. And it runs the wrong way for the mechanism in the literature,
which predicts more attention to later, lower regions and so rising rather than
falling accuracy.

A candidate explanation comes from the method's own authors, who report that a
steered model keeps its default ``panel 1, 2, 3\ldots'' narrative numbering while
the content follows the steer, and suggest that visual grounding and narrative
sequencing are functionally separate \citep{gandikota2026gaze}. If the model has a standing
preference for describing the first panel, then steering to panel 1 pushes with
that preference and steering to panel 6 against it, and accuracy should fall with
index for any head set, which is what we see. We have not tested this and offer
it as a hypothesis; the measurement that would test it, steering under a shuffled
panel order so that narrative position and visual position come apart, we did not
run. The audio arms are a separate matter, their gradient being small at 0.04
with its lowest point at the sixth segment, the one our encoder window truncates.

The horizontal decline
does not touch the vertical cells the transfer claim rests on, nor the audio
arms. We did not run the same check on the Bunny layouts, so we flag it as a
caution on the horizontal vision numbers generally rather than as a measured
property of all of them.

\begin{figure}[t]\centering
\includegraphics[width=\columnwidth]{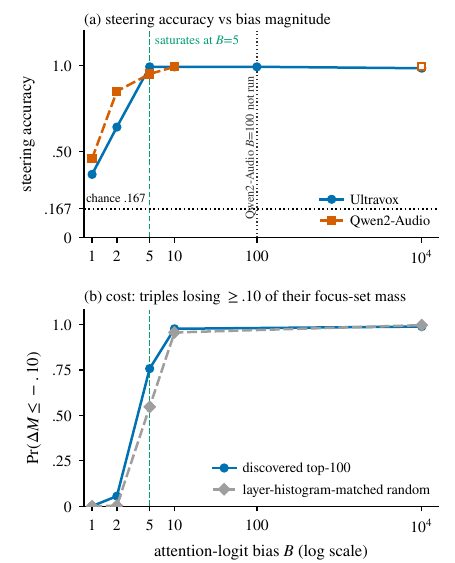}
\caption{Steering accuracy against bias strength, and where the target's gained
attention comes from. The effect saturates by $B{=}5$, so the large value used
for comparability with published work buys no accuracy, and at $B{=}5$ most
of the gained attention already comes from the prompt and the attention sink
rather than from the other segments.}
\label{fig:dose}
\end{figure}

\section{What our instruments cost}
\label{sec:cost}

An intervention can be fully effective on its target metric while doing most of
its work somewhere the metric does not look, and a control can be a distribution
wide enough to reverse the difference it is measuring. Both are true of the
instruments in this paper, and we measure them here rather than assert their
soundness.

\paragraph{Where the steered attention comes from.}
An additive pre-softmax bias does not create attention; it moves it, and
\citet{skop2026} show that query-space steering, which likewise shifts attention
logits across keys, pulls attention away from the small set of keys each head
actually relies on. We take their measurement. For
each head, its \emph{focus set} is the smallest group of keys holding 80\% of the
attention it pays in the unedited model, and we record how much of that survives
the edit. The thresholds below are theirs, so our figures sit beside the 31, 22,
and 10\% they report for their own steering method on one task. Everything in this paragraph is Ultravox at $K{=}100$,
across all 12{,}000 combinations of 100 heads, 20 strips, and 6 target segments;
we did not run it on the other two models.

The bias strength matters, and the place to read the result is where our own
effect saturates. Ultravox steers at 0.987 by $B{=}5$ against 0.990 at
$B{=}10^4$, so $B{=}5$ is the operating point and the larger value we use for
comparability with published work buys no accuracy. Figure~\ref{fig:dose} shows
both the dose response and where the gained attention is taken from.

At $B{=}5$, 75.8\% of combinations lose at least a tenth of their focus-set mass.
Of the attention the target segment gains, 0.173 is taken from the other five
segments, 0.416 from the prompt tokens, and 0.412 from the attention sink. That is roughly one part
intended redistribution to 4.8 parts taken from elsewhere. At $B{=}10^4$ it
degenerates further, to 0.069 from the other segments against 0.368 from the
prompt and 0.563 from the sink.

This constrains how Section~\ref{sec:borrowed} should be read, and we want to be
exact about how far. It does \emph{not} follow that the heads are sink-suppression
sites rather than segment trackers: the head set is selected by a
segment-conditional criterion computed on the clean model, with no intervention
at all, so whatever the bias does at $B{=}10^4$, the structure the ranking found
was there before anything was edited. What does follow is that the intervention
is a blunter instrument than its accuracy suggests. The natural reading, that the
bias reallocates attention from the unwanted segments to the wanted one, is true
at $B{=}2$, where 0.821 of the gain does come from other segments. It is not what
happens at the strength that works.

One experiment would settle what the edit is doing and we have not run it:
biasing the attention sink alone, with the segment spans untouched, to see how
much of the steering survives. We flag it as the obvious next measurement rather
than leave a reader to notice its absence.

\paragraph{Our own heads are disrupted more, not less.}
At $B{=}5$ the discovered heads lose a tenth of their focus mass in 75.8\% of
triples against 54.7\% for a layer-matched random set; the two converge by
$B{=}10$. One candidate explanation is in the same measurement: the discovered
heads have larger focus sets (9.5 keys against 5.2), so a fixed logit shift
displaces more absolute mass. We have not tested that and offer it as a
hypothesis rather than an account.

\paragraph{A matched random control is a distribution, not a number.}
Layer-histogram-matched random head sets are the standard negative control for
this kind of claim, and at the set sizes used here they are far less stable than
their use implies. Five draws of one construction, matched in size, in per-layer
counts, and in exclusions, span 0.008 to 0.650 of steering accuracy on Ultravox
at 29 heads. Five draws on Qwen2-Audio at the comparable size of 34 heads span
0.133 to 0.158. The two are a factor of 26 apart in width, though the paragraph
below shows that widths are not comparable across set sizes within a model
either, so we read this as two models behaving differently rather than as a
measured ratio. A difference quoted against a single draw can carry either sign:
on Ultravox the drawn control ranged from far below the subset it controls to far
above it. Five draws also understate a range, since the smallest and largest of
five are a conservative estimate of the extremes, so the instability is if
anything greater than these numbers show.

We can say when it matters but not yet why it varies. Doubling the set size
halves the spread on Ultravox (0.642 to 0.258 from 29 to 71 heads) and more than
doubles it on Qwen2-Audio (0.025 to 0.158 from 34 to 66), so it is not simply a
matter of averaging over more heads, and with three models we do not attempt an
account. What is actionable is the comparison rather than the cause: where the
effect is an order of magnitude above the spread, as for the results in
Section~\ref{sec:borrowed}, which draw was taken does not matter; where it is
comparable, as in a subset-versus-control difference, the draw decides the sign.

We draw two practices from this and follow them throughout. Layer-matched random
controls are reported as distributions over several draws, all of them printed
in Appendix~\ref{app:sets}, except the single matched set in the rerouting
measurement above. And a difference between a fixed head
set and a random control carries no bootstrap interval, because resampling items
measures the wrong variance. The dominant term is which heads were drawn.
Intervals appear in this paper only between two fixed sets on the same items.

We note this for the wider genre rather than for ourselves. ``The identified
units beat a matched random baseline'' is now a standard form of evidence in
interpretability \citep{saerandom2026,unittests2026,core2026heads,modalityconflict2026},
and on this evidence a single draw is not enough to support it. We looked for
prior measurements of how far such controls vary between draws and found none, so
we offer ours as a first estimate rather than as a correction to a known figure.

\begin{figure*}[t]\centering
\includegraphics[width=\textwidth]{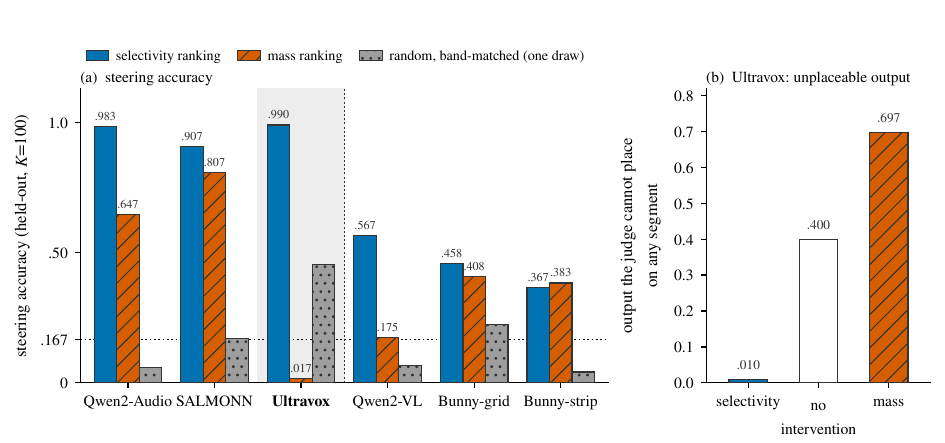}
\caption{The two head sets read two ways. \textbf{(a)} Steering accuracy across
six arms, beside one band-matched random draw per arm. Each draw spans the
selectivity-ranked hundred's layers; on Ultravox only 32 of the mass-ranked
hundred's heads lie in that span, so there the draw is not a control for the mass
ranking. \textbf{(b)} Ultravox read for unattributability on the same items:
the judge cannot place the output on any segment on 69.7\% of trials under the
mass-ranked heads, against 40.0\% with no intervention and 1.0\% under the
selectivity-ranked heads. The two head sets sit at different depths, so the
difference is not yet attributable to the score alone.}
\label{fig:failure}
\end{figure*}

\section{Where an attention-mass ranking parts from the tracking heads}
\label{sec:critique}

Section~\ref{sec:borrowed} showed that the tracking heads are inherited. A tempting
consequence is that they should be invisible to a ranking by attention mass,
because text pretraining never had audio to attend to. The data do not support
that version, and inheritance alone does not predict how a mass ranking treats
these heads, since the heads it returns come from the same pretrained model.

The mass ranking scores a head by its attention on the segment asked about
(Section~\ref{sec:method}), and every steering figure for it uses that score. Part
of what follows also uses a simpler quantity, a head's attention on the whole
audio block, which we call its \emph{audio attention} to keep the two apart. The
two order heads similarly, with Spearman $\rho = 0.844$, $0.932$ and $0.940$ and
70, 77 and 83 of their top 100 heads in common, but not identically.

The tracking heads do not attend little to the audio. Their medians sit at the
88th, 92nd, and 62nd percentile of their model's audio attention, at or above the
typical head on all three models. Whether a ranking returns them depends on where
its cut falls, which Table~\ref{tab:cutoff} shows for both quantities.

\begin{table}[t]\centering\small\setlength{\tabcolsep}{4pt}
\begin{tabular}{lccccc}
\toprule
 & & \multicolumn{2}{c}{audio} & \multicolumn{2}{c}{mass ranking's} \\
 & & \multicolumn{2}{c}{attention} & \multicolumn{2}{c}{score} \\
\cmidrule(lr){3-4}\cmidrule(lr){5-6}
model & cut & median & below & median & below \\
\midrule
Qwen2-Audio & 90.2 & 87.7 & 52 & 94.4 & 31 \\
SALMONN     & 93.8 & 91.6 & 64 & 91.8 & 63 \\
Ultravox    & 90.2 & 62.3 & 93 & 50.5 & 96 \\
\bottomrule
\end{tabular}
\caption{Where a top-100 cut falls in each model's own distribution, as a
percentile, against the median of the selectivity-ranked hundred and the number
of those hundred heads below the cut. The cut falls at a different percentile in
the 1024-head models than in 1600-head SALMONN, so it is given per model.}
\label{tab:cutoff}
\end{table}

In audio attention the median tracking head sits below the cut on every model,
and most tracking heads fall below it. On the mass ranking's own score the median
sits below the cut on SALMONN and Ultravox, on Ultravox at the typical head, and
above the cut on Qwen2-Audio. Stated as a set, the mass ranking shares 69 of its
top 100 heads with the selectivity ranking in Qwen2-Audio, 37 in SALMONN, and 4
in Ultravox, so it misses under a third of the tracking heads on one model and
nearly all of them on another.

Ranked across all heads, the orderings are only weakly related in every model.
Spearman $\rho$ between selectivity and audio attention is $+0.168$, $+0.258$ and
$+0.025$, and between selectivity and the mass ranking's score $+0.356$,
$+0.210$ and $-0.100$. A strong relationship would have made the two rankings
interchangeable. A weak one allows the tops of the two rankings to come apart
without requiring it, and the overlaps above show both outcomes. The mass-ranked
hundred carries 1.2 to 2.9 times the audio attention of the selectivity-ranked
hundred, and steers less accurately than it on every audio model.

We tried to turn that into a claim about \emph{why}, and it did not hold. The
natural explanation is that mass ranking returns the heads whose perturbation is
most disruptive, and two comparisons appear to support it: splitting our own
hundred on Ultravox by audio attention gives 0.850 for the lower half against
0.647 for the upper half, and the fifty heads highest in audio attention within
the tracking heads' layer span, excluding the selectivity-ranked hundred, collapse
to 0.017. Both are confounded. Matched on
depth, the two halves steer at 0.687 and 0.677 despite one carrying 2.5 times the
audio attention of the other, so the split was measuring depth rather than
attention. And the set that collapses sits at mean layer 12.9, where almost none
of the tracking heads live. We report the divergence between the rankings, which
holds, and not an account of what the mass ranking selects instead, which we do
not have.

\paragraph{Consequences under intervention.}
On Qwen2-Audio and SALMONN the mass-ranked heads do steer, at 0.647 and 0.807
against band-matched random controls of 0.060 and 0.170. On Ultravox they reach
0.017, below the 0.080 the model reaches with no intervention at all, where the
selectivity ranking reaches 0.990.

The natural way to describe that gap, as the published score steering below
random, is not available to us, and we say so because we tried it first. Ten
random hundreds drawn to the selectivity-ranked hundred's per-layer counts,
excluding it, span 0.007 to 0.663 in steering accuracy with a mean of 0.250, and
two of the ten land at or below 0.017, one of them exactly on it. Those draws
could not settle the question in any case. They sit where the tracking heads sit,
at mean layer 22.8, while the mass-ranked hundred sits at mean layer 6.2, with 84
of its heads in layers the draws never sample. The control the claim needs is a
random hundred at the mass-ranked heads' own depth. We have not run it, and make
no claim in either direction.

A ranking whose top heads make a task worse under intervention is not new.
\citet{writerscancellers2026} split function-vector heads into writers and
cancellers by sign-preserving path patching, and find that a magnitude-ranked
ablation baseline preferentially recovers the cancellers on their hierarchical
task, and that removing the cancellers raises the correct-label logit difference
in all six of their main cells. Their score is
causal-effect magnitude rather than attention mass, their remedy is a sign rather
than a per-head normaliser, and they lesion rather than steer to a chosen target.
The shape of the finding is theirs before it is ours.

\paragraph{The two failing sets fail differently.}
Steering accuracy hides a distinction. A head set can fail by making the model
describe the wrong speaker, or by making it describe nothing the judge can place
at all, and both score the same. Table~\ref{tab:failmode} separates them, and
Figure~\ref{fig:failure} shows the same runs read both ways.

\begin{table}[t]\centering\small
\begin{tabular}{lcccc}
\toprule
 & \multicolumn{2}{c}{selectivity ranking} & \multicolumn{2}{c}{mass ranking} \\
\cmidrule(lr){2-3}\cmidrule(lr){4-5}
model & names & cannot & names & cannot \\
      & target & place  & target & place \\
\midrule
Qwen2-Audio & 0.983 & 0.013 & 0.647 & 0.143 \\
SALMONN     & 0.907 & 0.080 & 0.807 & 0.153 \\
Ultravox    & 0.990 & 0.010 & 0.017 & \textbf{0.697} \\
\bottomrule
\end{tabular}
\caption{Proportion of trials on which the model names the target segment, and on
which the judge can place the description on no segment at all. The two need not
sum to 1: the remainder names a segment, and the wrong one.}
\label{tab:failmode}
\end{table}

On Qwen2-Audio and SALMONN the two rankings fail in much the same way, with a
little more unplaceable output under the mass ranking than under the selectivity
ranking. Ultravox is the exception, and a large one. There the mass-ranked heads
leave output the judge cannot place on any segment on 69.7\% of trials, against
40.0\% with no intervention and 1.0\% under the selectivity ranking: one head set
directs the description and the other mostly leaves it unplaceable. The two
hundreds sit about seventeen layers apart, so this difference, like the accuracy
gap, belongs to the two rankings as they come out and cannot yet be assigned to
the score rather than to depth.

We can measure that and cannot explain it. The obvious account, that these are
the highest-mass heads and perturbing them is maximally damaging, is not
supported by the depth-matched halves above at the tracking heads' depth, and is
untested at the mass-ranked heads' own depth. It is also close to untestable at this
set size: the mass-ranked hundred is by construction the hundred scoring highest
on its score, so no control can hold that score constant while varying whether a
set is the mass-ranked one. Depth can be held constant, and that is the control
this result lacks.

An attention measure doing worse than simple alternatives has also been reported
at the level of maps.
\citet{sadanandan2026grounding} report that, for one medical VLM on one dataset, true-box attention coverage is the
lowest of four conditions, below shifted and randomly placed boxes, and that the
ordering holds across a nine-group layer sweep. They average attention over heads
and layers rather than ranking or selecting heads, they occlude rather than
steer, and they propose no replacement score. Ours is a head-level relative, on
one arm of six and with a weaker control than theirs: a head-selection score whose
heads, under intervention, leave the output unattributable on most trials, beside
a per-head normalised variant of the same score that reaches near ceiling on the
same items.

\paragraph{Scope.}
Everything specific to Ultravox in this section is one arm of six. On the other
five, three of them vision models, the mass ranking steers above a single
band-matched random draw, on Qwen2-VL only just above chance. Across all six, a
weaker claim holds on four: ranking by selectivity rather than by mass selects
better-steering heads, each comparison run on the same items so the two rankings
can be differenced directly, with those four intervals excluding zero and the two
Bunny layouts indistinguishable. The arm where the mass ranking fails is
also the arm where the two rankings diverge most, sharing 4 of 100 heads against
37 and 69. We report that coincidence and decline to build on it, because with
three audio models and one failure the failing arm is extreme on five counts at
once. It has the lowest modality attention of the three, it is the only one with
a fully frozen language backbone, its random controls at the unshared-subset size
vary most widely (Section~\ref{sec:cost}), its two rankings overlap least, and its mass-ranked
heads sit far shallower, at mean layer 6.2 against 23.2 and 20.8. Nothing here
distinguishes which of the five matters.

\paragraph{Norm weighting partly lifts accuracy and leaves the failure mode.}
Both rankings read attention after the softmax. Weighting each head's attention
by the norm of what it writes, which reverses the classic finding that BERT
attends to \texttt{[SEP]} \citep{kobayashi2020attention}, lifts the mass ranking
on Ultravox from 0.017 to 0.160 under $\lVert v \rVert$ and 0.203 under
$\lVert W_O v \rVert$, nine- to twelvefold, while its unattributable rate barely
moves, to 65.0\% and 63.3\%. The selectivity ranking barely moves either, from
0.990 to 0.970, and keeps 99 of its top 100 heads.

So a one-line change to how attention is measured accounts for part of the
accuracy collapse and little of the unplaceable output. At 0.203 the corrected
mass ranking recovers under a fifth of the distance to the selectivity
ranking.

\paragraph{Why the selection step is worth this much attention.}
A head-selection score is not only an explanation. Downstream methods crop, mask,
and reweight attention on the heads such scores return
\citep{zhangma2026ibs,maskcd2025,kang2025sink}, and whatever a selection step
gets wrong is inherited by all of them. We measure steering only, and make no
claim about how those methods would fare under one ranking or the other.

\section{A published null, and a preprocessing step that could explain it}
\label{sec:null}

The seed method reports its head set absent in two model families, with Bunny at
8.3\% steering accuracy against 16.7\% chance. That null is load-bearing: it is
the evidence that gaze heads form in some architectures and not others, and the
authors read it as pointing at the vision encoder. So we tried to reproduce it.

We keep this finding separate from the instrument critique in
Section~\ref{sec:critique}, as our preregistration requires. The reason is worth
stating rather than implying: a failure to reproduce a null and a criticism of a
score are different kinds of evidence, and combining them would let a weakness in
either borrow support from the other. Nothing in this section depends on
Section~\ref{sec:critique}, and nothing there depends on this.

We do not reproduce the null, on either layout we build. With the six panels in a
grid, Bunny steers at 0.408 under the mass ranking and 0.458 under the
selectivity ranking, against 0.225 for a band-matched random control. With them
in a 6:1 strip, the figures are 0.383 and 0.367 against 0.042. All are far above
the 16.7\% chance their 8.3\% sits below.

Their own discovery statistic points the same way, though weakly, and we report
it as weak. Recomputing it on our data gives real-versus-shuffled concentrations
of 1.513 on the grid and 1.733 on the strip, where 1.0 would mean the ranking is
indistinguishable from one built with the panel labels shuffled. Those are above
1.0 and not far above it; on Qwen2-VL the same statistic reaches values several
times larger. We take them as consistent with a head set being present in Bunny,
not as strong evidence of one.

\paragraph{One explanation ruled out.}
Their score reads the final prompt token. Measured during generation instead,
head concentration is higher on every arm where we measured both, by 1.56 to 2.57
times on the audio arms and Qwen2-VL and from 1.540 to 2.122 on Bunny, so that
token understates it. If prefill selected materially different heads, the null would be an
artefact of when they looked. It does not: the sets selected at the two
measurement points overlap on 72 of 100, against a layer-matched null of 22.4.

\paragraph{One that takes the arm to chance.}
Their Appendix E.3 describes replacing a centre crop with a forced square resize
for fixed-resolution model families, naming Ovis, InternVL and both LLaVA
variants. Bunny is not among them.
Centre-cropping the 6:1 strip keeps only the second and third panels, each about
half visible, and four of six panels leave the image entirely. Applying that crop
to the strip arm takes steering from 0.367 to 0.142 under the selectivity ranking
and from 0.383 to 0.158 under the mass ranking, and erases discovery:
real-versus-shuffled concentration falls to 0.998 and 0.987, which is to say to
the value a shuffled ranking would give.

\paragraph{What this does and does not establish.}
It establishes that a preprocessing step named in the original's own appendix as
harmful, and not applied to this family there, is sufficient to take a working
arm to chance and destroy its head structure. It does not establish that this
step was applied: a family's absence from a list of fixes is not evidence that
the unfixed behaviour occurred, and we cannot inspect their pipeline. Nor do we reproduce their
8.3\%, since their number sits below chance and our cropped arm sits at it.

The material is not an explanation either, because it is theirs. We take 150
items from the released \texttt{baulab/openai-comic-strips}, tile each item's six
panels into the two layouts, and score answers against the captions shipped with
the dataset. Two differences from their setup remain and neither concerns the
images themselves: we use 150 of the 500 items, and we resize panels to 448
pixels square before tiling, where the released panels are about 1024 wide.

\section{Conclusion}
\label{sec:conclusion}

The machinery an audio model uses to track who is speaking is machinery its
language backbone already had. Heads chosen on a model's text-only backbone, or for Qwen2-Audio on a same-family
text model,
without any audio entering the selection, steer the audio model to a chosen
speaker on all three models we test, and the subset shared between the audio and
text head sets is causally sufficient for the behaviour rather than merely
co-located with it. Whether being \emph{shared} is what makes those heads work is
a further question, and one our controls do not settle on any of the three
models; we report that boundary wherever the claim appears.

A second finding concerns method: ranking heads by how much attention they place
on the region asked about retrieves a partly different set from ranking them by
how selectively they place it, in all three audio models, by an amount that
ranges from 69 shared heads of 100 to 4. On Ultravox, one of six arms, the difference is severe enough that the two head
sets fail differently under the same intervention: the mass-ranked heads leave
output the judge cannot place on any segment on 69.7\% of trials, against 40.0\%
with no intervention, while the selectivity ranking names the target on 99.0\%.
The two head sets also sit at very different depths, and we have no control that
separates the score from that.

Two findings about our own instruments generalise further than the rest of this
paper, and we would rather they were used than admired. First, an attention-logit
edit can be fully effective on its target metric while doing most of its work
elsewhere: at the strength where ours saturates, the target segment takes
0.173 of its gained attention from the other segments, 0.416 from the prompt, and
0.412 from the attention sink. Anyone reporting such an edit can run this
decomposition cheaply, and we think they should.

Second, ``the identified units beat a matched random baseline'' has become a
standard form of evidence in interpretability, and a single random draw is not
enough to support it. Five draws of one identically constructed control span
0.642 of accuracy on one model and 0.025 on another. In our own analysis a
difference measured against a single draw carried the wrong sign until we drew
four more. Controls of this kind should be reported as distributions, and
differences against them should not carry intervals computed by resampling items,
which measures the wrong thing.

\section*{Limitations}
\label{sec:limitations}

In rough order of how much they threaten the results.

\paragraph{4-bit quantisation.} The three
audio models and Qwen2-VL run in 4-bit NF4 with bf16 compute; Bunny runs in
fp16. \citet{pahq2025} measure what low precision does to circuit discovery
and report that 4-bit collapses ACDC's edge-recovery AUC-ROC on GPT-2 to
0.61 against 0.96 at 16-bit, concluding it is ``almost impossible to use 4-bit
quantization''. That is a direct threat to any head ranking computed at this
precision, and we do not have a re-ranking at bf16 to set against it. Three
things bound the threat without removing it: their evidence is a single ablation
on a small model and one task, their low-precision setting quantises activations
as well as weights rather than NF4 weights with bf16 compute, and their measured
object is ACDC edge recovery rather than attention-mass head ranking. None of that is a substitute for
the experiment. A bf16 re-ranking,
reported as rank correlation and top-100 overlap against our split-half band of
91--97, is the first thing we would run with more compute.

\paragraph{The audio arms have no layout control.} Section~\ref{sec:geometry}
separates segment tracking from position tracking by moving content within the
token sequence, which we did not do for audio, where each segment's tokens are
contiguous in time. For audio, the entire
positional evidence is that per-target accuracy on Ultravox is flat (spread 0.04),
at a near-ceiling accuracy that leaves little room to vary. Flatness
is weaker than transfer, and the vision result is evidence by analogy.

\paragraph{Our intervention satisfies a known dormancy condition by
construction.} A pre-softmax additive bias sets attention logits to values no
clean forward pass produces, which is the off-distribution regime
\citet{makelov2024illusion} identify as capable of activating dormant pathways;
the reply of \citet{wu2024rebuttal} argues the concern is narrower than stated. We
take no position, and note that our steering results are therefore evidence about
what the model \emph{can} be made to do through these heads, not necessarily
about what it does unprompted. The concern is sharpest at $B{=}10^4$, which drives
attention close to one-hot and is the regime \citet{zhang2024pasta} warn against
for their own attention reweighting; Section~\ref{sec:cost} is why we report
$B{=}5$ alongside it, where the effect is already saturated.

\paragraph{The Ultravox failure has no control at its own depth.} The mass-ranked
hundred on Ultravox sits at mean layer 6.2, and every random control we ran on
that model was drawn at the selectivity-ranked hundred's layers. Its
unattributable rate is compared against the selectivity ranking and against no
intervention, both on the same items, but the selectivity-ranked hundred sits at
a different depth (mean layer 22.8 against 6.2). A random hundred at the mass-ranked heads' own
depth is the comparison that would say whether the failure belongs to the heads
the score chose or to where they sit, and we have not run it.

\paragraph{Head-set overlap is not identity of function.}
Section~\ref{sec:geometry} shows the mechanism survives a change of token
geometry; it does not show the computation performed is the same one. The
set-level intervention in Section~\ref{sec:borrowed} leaves open whether each
shared head plays the same role in both settings, a question close to the one
\citet{computationalunit2026} pose about when a head is a unit of computation.

\paragraph{Scale.} Three audio models, one vision model family for the layout
control, one corpus per modality, 50 items by six targets per confirmatory
steering cell and 20 for the subset comparisons and the dose sweep. Several
findings are explicitly one-of-three or one-of-six and are marked as such
wherever they appear.

\paragraph{One of six segments is only partly observed.} The audio strips are
33.5 seconds and the encoders' window is 30, so the sixth segment is truncated at
about 30\% of its duration. Every audio number here is a mean over six diagonal
entries of which one is computed on a stub. Rebuilding the testbed at four-second
segments would remove this and changes the corpus, so we report it rather than
patch it.

\paragraph{Head rankings are static.} Both scores are time-averaged over the
prompt. \citet{dynamicheads2026} argue retrieval-head identity varies by
generation timestep. A split-by-generation-position stability measurement is
registered and not run.

\paragraph{The horizontal vision arm has an unexplained gradient.} On the
horizontal Qwen2-VL layout our steering declines monotonically with panel index,
from 0.70 to 0.35 ($\rho = -0.95$). Every cell we measured declines with index,
including a random control at $-0.53$, so the gradient sits in the model or the
task rather than in the head set; what distinguishes this arm is the size of it.
Section~\ref{sec:geometry} offers an untested explanation from the original
authors. We did not run the check on the Bunny layouts, so we treat it as a
caution on the horizontal vision numbers rather than a measured property of all
of them.

\paragraph{The non-reproduction differs from the original in setup, not in data.}
Section~\ref{sec:null} uses the released comic-strip corpus, so a difference in
the material is not an explanation for that section, but it uses 150 of the 500
items and resizes panels to 448 pixels square before tiling. We have also never run the original's
code end to end: its release covers one model family and contains no reference to
the families whose nulls we examine, so those nulls came from code that was not
released.

\paragraph{We do not offer a better instrument.} The selectivity score is used to
make a comparison, not proposed as a replacement. On Ultravox, dropping the
half of our own hundred highest in audio attention and steering with the remaining 50 heads
reaches 0.850, against 0.643 for our plain top 50 on the same held-out items. Our
own ranking is not the best selection rule even among rules we have tried. The
per-head normalisation and the layer-matched control have both been used before
\citep{kang2025sink, shengfu2026matched}, correctly applied here, and
neither is a contribution.

\section*{Acknowledgements}

This work was carried out with substantial assistance from a large language
model, used for code, analysis, literature search and drafting throughout, and
disclosed here under the ACL policy on generative AI in authorship. The author
is solely responsible for the claims made.

Every experimental result reported here comes from runs on the author's own
hardware. Figures and tables are generated from a results ledger built directly
from those run outputs rather than transcribed, and a guard script fails the
build on any quantity appearing in the text that cannot be traced to the ledger.
Literature identified with model assistance was retrieved and read before being
cited.

\begingroup\makeatletter
\def\@listi{\leftmargin\leftmargini \parsep 1pt \topsep 2pt \itemsep 1pt}\let\@listI\@listi
\makeatother
\bibliography{main}
\endgroup

\clearpage
\onecolumn
\raggedbottom
\appendix
\section{Random control draws}
\label{app:sets}

Section~\ref{sec:cost} measures that a random head control at these set sizes is a
distribution rather than a point. This appendix prints the individual draws behind
the controls the subset comparisons of Section~\ref{sec:borrowed} rest on, and the
audio controls at $K{=}100$ in Sections~\ref{sec:family} and~\ref{sec:critique}.
Where the text gives one of these controls as a single number, it is the mean of
the draws printed here, except the band-matched figures at $K{=}100$, which are the
first draw (Section~A.3); other random controls in the text are single draws.
Entries are steering accuracy on held-out strips, with chance at $1/6$, except
where a column is marked as the unattributable rate.

\subsection{Draws from inside the discovered hundred}

These are the controls in Table~\ref{tab:sufficiency}. Each draw is taken uniformly
at random from the discovered top-100 itself, at the tested subset's size, the
exception to the usual pool stated in Section~\ref{sec:method}. Each draw is 20
held-out strips by 6 targets, or 120 trials. The share of each draw that is also
in the backbone's text top-100 is given beside it, because it sets the contrast:
at the shared subset's size a draw cannot avoid being mostly shared, and at the
unshared subset's size there is no such floor.

\begin{table}[H]\centering\small
\begin{tabular}{lcccccc}
\toprule
model & heads & draw 1 & draw 2 & draw 3 & subset & \% of each draw shared \\
\midrule
\multicolumn{7}{l}{\emph{size of the shared part: narrow contrast}} \\
Qwen2-Audio & 66 & 0.958 & 0.933 & 0.917 & 0.975 & 68, 73, 65 \\
Ultravox    & 71 & 0.842 & 0.933 & 0.975 & 0.950 & 69, 70, 72 \\
SALMONN     & 74 & 0.842 & 0.792 & 0.958 & 0.775 & 72, 70, 72 \\
\midrule
\multicolumn{7}{l}{\emph{size of the unshared part: wide contrast}} \\
Qwen2-Audio & 34 & 0.358 & 0.683 & 0.892 & 0.275 & 62, 65, 68 \\
Ultravox    & 29 & 0.667 & 0.417 & 0.250 & 0.383 & 69, 69, 69 \\
SALMONN     & 26 & 0.342 & 0.517 & 0.675 & 0.592 & 69, 73, 81 \\
\bottomrule
\end{tabular}
\caption{The three draws behind each random figure in Table~\ref{tab:sufficiency},
beside the fixed subset they control: the shared part in the upper block and the
unshared part in the lower. The means quoted in Table~\ref{tab:sufficiency} are
0.936, 0.917 and 0.864 above and 0.644, 0.444 and 0.511 below.}
\label{tab:app-inside}
\end{table}

Three subsets fall outside all three of their draws: Qwen2-Audio's shared part,
above its draws, and Qwen2-Audio's unshared part and SALMONN's shared part, both
below theirs. A subset no different from its draws would fall outside all three
half the time, so three of six is what chance alone predicts, and no single row
is decisive. Only the wide contrast can separate being shared with the backbone
from being in the discovered hundred at all, because at the narrow contrast the
draws are themselves about 70\% shared (Section~\ref{sec:borrowed}). Among the
wide contrasts, Qwen2-Audio's unshared part, at 0.275 against draws of 0.358, 0.683 and 0.892, is
the strongest case, and by the same arithmetic a subset no different from its
draws would sit below all three a quarter of the time. No model separates the two
explanations.

\subsection{Layer-matched draws from outside the hundred}

These are the controls behind the sufficiency claim and the diamonds in
Figure~\ref{fig:sufficiency}. Each draw reproduces the subset's per-layer counts
and is taken from outside the discovered hundred, 120 trials per draw.

\begin{table}[H]\centering\small
\begin{tabular}{lccccccc}
\toprule
model & heads & draw 1 & draw 2 & draw 3 & draw 4 & draw 5 & subset \\
\midrule
\multicolumn{8}{l}{\emph{controls for the shared part}} \\
Qwen2-Audio & 66 & 0.075 & 0.183 & 0.100 & 0.233 & 0.167 & 0.975 \\
Ultravox    & 71 & 0.150 & 0.217 & 0.408 & 0.200 & 0.208 & 0.950 \\
SALMONN     & 74 & 0.192 & 0.242 & 0.067 & 0.067 & 0.450 & 0.775 \\
\midrule
\multicolumn{8}{l}{\emph{controls for the unshared part}} \\
Qwen2-Audio & 34 & 0.142 & 0.158 & 0.133 & 0.158 & 0.150 & 0.275 \\
Ultravox    & 29 & 0.650 & 0.008 & 0.158 & 0.325 & 0.383 & 0.383 \\
SALMONN     & 26 & 0.383 & 0.033 & 0.308 & 0.033 & 0.083 & 0.592 \\
\bottomrule
\end{tabular}
\caption{Five draws per control, beside the fixed subset each one controls. The
shared part exceeds all five of its draws on every model. The unshared part does
so on Qwen2-Audio and SALMONN and not on Ultravox, where it is above three draws,
level with a fourth and below the fifth. That Ultravox row, spanning 0.642, is the
widest of these controls and the first example in Section~\ref{sec:cost}'s
discussion of control spread.}
\label{tab:app-outside}
\end{table}

\subsection{Controls at \texorpdfstring{$K{=}100$}{K=100}}

Each draw at this size is 50 held-out strips by 6 targets, or 300 trials.

\paragraph{Band-matched.} The audio band-matched figures at this size in
Sections~\ref{sec:family} and~\ref{sec:critique} are draw 1 of the two below.
Each draw is taken uniformly from the layers the selectivity-ranked hundred
spans, excluding that hundred. On Qwen2-Audio the mass-ranked hundred has 98 of
its heads inside that span and on SALMONN 73; on Ultravox only 32, so there these
draws are not a control for the mass ranking. The second draw changes no
comparison the text makes: the mass ranking stays above both draws on Qwen2-Audio
and SALMONN.

\begin{table}[H]\centering\small
\begin{tabular}{lcccccc}
\toprule
 & \multicolumn{2}{c}{accuracy} & \multicolumn{2}{c}{unattributable} & \multicolumn{2}{c}{mass ranking} \\
\cmidrule(lr){2-3}\cmidrule(lr){4-5}\cmidrule(lr){6-7}
model & draw 1 & draw 2 & draw 1 & draw 2 & accuracy & unattributable \\
\midrule
Qwen2-Audio & 0.060 & 0.153 & 0.127 & 0.107 & 0.647 & 0.143 \\
SALMONN     & 0.170 & 0.187 & 0.497 & 0.413 & 0.807 & 0.153 \\
Ultravox    & 0.453 & 0.527 & 0.393 & 0.277 & 0.017 & 0.697 \\
\bottomrule
\end{tabular}
\caption{Both band-matched draws at $K{=}100$ on the three audio models, beside
the mass ranking on the same items.}
\label{tab:app-band}
\end{table}

\paragraph{Ten layer-matched draws on Ultravox.} These draws copy the per-layer
counts of the selectivity-ranked hundred, which spans layers 10 to 30 at mean
layer 22.8, and exclude it. They are therefore a layer-matched control for that
hundred, which steers at 0.990 against a highest draw of 0.663. They are not a
control for the mass-ranked hundred, which spans layers 0 to 18 at mean layer 6.2
(Section~\ref{sec:critique}).

\begin{table}[H]\centering\small
\begin{tabular}{lcc}
\toprule
head set & accuracy & unattributable \\
\midrule
draw 1  & 0.393 & 0.223 \\
draw 2  & 0.283 & 0.240 \\
draw 3  & 0.663 & 0.120 \\
draw 4  & 0.200 & 0.143 \\
draw 5  & 0.147 & 0.257 \\
draw 6  & 0.177 & 0.450 \\
draw 7  & 0.017 & 0.287 \\
draw 8  & 0.007 & 0.373 \\
draw 9  & 0.073 & 0.233 \\
draw 10 & 0.543 & 0.187 \\
\midrule
mean of the ten & 0.250 & 0.251 \\
\midrule
selectivity-ranked hundred & 0.990 & 0.010 \\
\bottomrule
\end{tabular}
\caption{Ten random hundreds on Ultravox at the selectivity-ranked hundred's
layers, with that hundred on the same items for reference.}
\label{tab:app-ten}
\end{table}

\end{document}